%% file: main.tex
\documentclass[10pt]{article}

\input{template/preamble}

\usepackage{lmodern}

\usepackage{graphicx}
\usepackage{times}

\begin{document}
\noindent

\bibliographystyle{ieeetr}

\title{Toward Fault Detection in Industrial Welding Processes \\with Deep Learning and Data Augmentation}

\authorname{J. Antony \textsuperscript{[a]}, Dr. F. Schlather \textsuperscript{[b]}, G. Safronov \textsuperscript{[b]}, M. Schmitz \textsuperscript{[b]}, Prof. Dr. K. V. Laerhoven \textsuperscript{[a]} }
\authoraddr{[a] University of Siegen, Germany; [b] BMW Group, Munich, Germany }


\maketitle


\abstract
With the rise of deep learning models in the field of computer vision, new possibilities for their application in industrial processes proves to return great benefits. Nevertheless, the actual fit of machine learning for highly standardised industrial processes is still under debate. This paper addresses the challenges on the industrial realization of the AI tools, considering the use case of Laser Beam Welding quality control as an example. We use object detection algorithms from the TensorFlow object detection API and adapt them to our use case using transfer learning. The baseline models we develop are used as benchmarks and evaluated and compared to models that undergo dataset scaling and hyperparameter tuning. We find that moderate scaling of the dataset via image augmentation leads to improvements in intersection over union (IoU)  and recall, whereas high levels of augmentation and scaling may lead to deterioration of results. Finally, we put our results into perspective of the underlying use case and evaluate their fit.




\section{Introduction}
\label{chp:intro}

As the fourth industrial revolution is enabling the digital transformation of industrial manufacturing processes, greater access to data encourages the development of Artificial Intelligence tools for the automation of manufacturing environments. Recent advances in digital image processing through computer vision and machine vision via deep learning (DL) make visual data processing more efficient and increasingly powerful. Thus, various industrial applications, particularly within the contact-less, non-destructive quality control and inspection, make use of these methods as e.g. in the automotive body shop.

A car body is manufactured by assembling a number of pre-pressed panels applying various material joining methods. Out of many joining processes, Laser Beam Welding (LBW) creates fast, continuous and long weld seams with notable mechanical advantages over the conventional resistance spot welding technique \cite{laser_adv_spot}. This LBW technology is widely used in the e-mobility sector, particularly for the manufacturing of the battery and electric motor components of electric cars \cite{laser_for_battery}. As usual in the automotive industry, reliable intelligent quality inspection systems are in place during the production process to reassure the desired outcome. Especially the monitoring of the LBW process involves diverse sensor units. We divide the overall welding process into three parts, as seen in Figure \ref{fig:data_sample}: pre-process, in-process and post-process. The pre-process can be used for the weld path planning; the in-process for the process prediction by identifying the characteristics of the molten pool and occurring splashes and the post-process for the evaluation of the finished surface by analysing the solid weld seam. Machine vision or image based inspection tools are a good fit for monitoring of all the mentioned stages of the LBW simultaneously \cite{stavridis2018quality}. 

In this paper, we conduct a study to evaluate the applicability of quality inspection via DL on the finished weld seams using the image data of the post-process, obtained directly from the LBW camera unit. As the process monitoring camera data shows multiple surface pores at certain instances, multiple classification and localization algorithms were utilized for this study. Also, the size of the identified surface pore determines the acceptance or rejection of a finished work piece, a size based evaluation is required for this use case. As part of the initial modelling, state-of-the-art (SOTA) object detection algorithms from the TensorFlow Object Detection API are adapted through transfer learning as baseline models. We then evaluate the baseline models and compare their performance on the available dataset. The best performing baseline models are further fine tuned using hyper parameter adjustments and dataset scaling by offline image augmentation methods. The performance of the models on scaled datasets are determined and the possible performance characteristics are identified.  

This paper is further structured as follows: We begin by reviewing related research literature in section \ref{sec:related-work}. We then discuss our methodology in section \ref{sec:methedology}. Afterwards, we describe the use case, the data understanding and the data preparation in sections \ref{sec:use-case}, \ref{sec:data-understanding} and \ref{sec:data-preparation}. Our augmentations to the computer vision algorithms is shown in section \ref{sec:modelling}. We then present the results of the study in section \ref{sec:results} and finally come to a conclusion and discuss further research potential in section \ref{sec:conclusion}.

\section{Related Work}
\label{sec:related-work}
Due to industrial quality standards, failure detection and quality control in production processes, especially through the implementation of AI tools are an important field of research in the area of Industry 4.0. There are many studies in the field of process monitoring on LBW process through data-driven processes. The pioneer research in LBW process control using Artificial Neural Networks was proposed in 2005 by \cite{Sound_ANN_2005}, where the generated sound during the welding is analyzed to predict the welding quality. A visual inspection method, based on anomaly detection on photographed images of welded pools, is proposed by the authors of \cite{gao2014prediction}, where a back propagation network classifies the failures. A DL based CNN decoder-encoder system for the LBW quality prediction, based on the optical microscopic (OM) images, is proposed by the authors of \cite{deep_keyhole_2020}. The proposed encoder network converts the OM images to feature maps and a GAN based decoder network could predict OM images, after taking few process parameters as input. Another visual inspection method using DL is proposed by the authors of \cite{8124091_CNN_defect}, where a CNN based classifier is used to identify the type of defects. This proposed network classified the defect into one of the four defined welding defects, with an accuracy of $95.33\%$, on the dataset. Following this research, the authors of \cite{Hairpin_CNN} developed CNN based DL monitoring system for the quality control of hairpins used in electric motors. This binary classifier model performed with an average accuracy of $74\%$ and average recall of $70\%$ on their datasets.  

Similarly, the area of general surface quality inspection is also an active field of research for AI tools implementations. A CNN based defect classification network called LEDNet, used in the production control of LED lights, has been developed by the authors of \cite{lin2019automated} in 2018. Another novel cascaded auto-encoder network for the segmentation and localization of the defects on the defect localization on metallic surfaces is developed by \cite{tao2018automatic}, where their model could generate a prediction mask over the failures. As the localization of the defects help the production operator to locate and to understand the defects more intuitively, object detection based models became more significant in the researches. In 2018, the authors of \cite{cha2018autonomous} proposed a near real-time defect detection method based on Faster-RCNN network for the inspection on concrete and steel surfaces. Their research gave an average precision of $87.8\%$ over the five defect classes. Following this in 2018, the authors of \cite{lin2018steel} performed another study by comparing the performances of Faster-RCNN and SSD object detection algorithms to detect and classify the defects. The authors also points out the inability of the YOLO network in detecting smaller objects. In another study in 2018, a real time object detection based on the modified version of YOLO network has been suggested by the authors of \cite{LI201876}. The modified YOLO network performed surface detection on six different types of classes with $97.55\%$ precision and $95.86\%$ recall rates on the datasets.   

There are multiple researches performed implementing AI tools in the classification, detection and segmentation tasks on identifying various defects in LBW applications and general surface quality control applications. The open-source communities in the AI developments accelerate the research activities in the realization of AI tools for industrial applications. The open-source object detection API framework from TensorFlow enables the easy implementation and comparison of various transfer learning networks on any new custom datasets. This research throws lights to such an industrial realization of object detection algorithms.  

\section{Methodology}
\label{sec:methedology}
This research is structured on the widespread data driven methodology for data mining, called the CRISP-DM (Cross-industry standard process for data mining) \cite{Marbn20121AD}. As explained in the chapter \ref{chp:intro}, baseline models are developed on the available dataset as initial study. The pre-trained models from the TensorFlow Object Detection API are modelled to fit the custom dataset using transfer learning. After developing the baseline models, the performance and inference time of the models are compared with the industrial demands of the use case. Based on this evaluation, the best performing model is identified and the same is further fine-tuned for performance improvements. The hyper-parameter optimization and dataset scaling through offline augmentation are performed for the fine tuning of the baseline model. The performance of the models upon dataset scaling is observed and possible reasons are evaluated. 

Initially, the use case and data have been studied and the optimum AI tools were selected based on the nature of the work. Upon that, the dataset has been prepared to match the workflow. Later the baseline model are developed, evaluated and inferences are made. Further fine tuning is performed and successive modelling, evaluation and performance analysis are done. Due to the research nature of this work, the final 'deployment in production' stage in CRISP-DM is replaced with inference step. For the baseline implementation the SOTA networks, the single stage object detection network based on the SSD Mobilenet \cite{10.1007/978-3-319-46448-0_2_SSD} and two stage object detection network based on the Faster-RCNN \cite{girshick2015fast_rcnn} are used. The trained models are evaluated using the COCO matrix \cite{lin2014microsoft} by comparing the precision (mAP) and recall (mAR) for various IoUs and object scales.     

\section{Use Case Understanding}
\label{sec:use-case}

The Use Case this study was aligned with is taken from the automotive body shop. To manufacture the complex structure of a car door, metal sheets that have been pressed into form are joined together by a variety of joining technologies, including LBW. In the laser cell, the separate parts are clamped in a fixture that assures correct placement. After fixation, a robot with a tactile laser welding tool follows the edge between the parts, welding the pieces together. The weld process is controlled by instruments and sensors attached to the laser welding tool, including an industrial camera for process observation. After the weld, the seam is inspected in an adjacent quality assurance cell. The assessment of the weld is necessary, as the laser welds are visible to the customer and due to corrosion protection.

In the observed business case, the quality of the weld has a direct impact on the overall quality of the doors. Due to the physical properties of the laser welding process, it is difficult to detect the various possible weld defects, which can occur. Many use-case related faults such as weld-throughs and inner pores are not observable during the welding process. Surface pores on the other hand are directly visible to the observation camera, but were hard to be automatically detected by a conventional computer vision approach due to the big feature variances, like size and form. Considering these constraints and the underlying business case, the evaluation of modern computer vision methods, especially DL, are one step in improving both the accuracy of the quality assessment as well as reduce the effort and time needed to evaluate each weld. 

\section{Data Understanding}
\label{sec:data-understanding}
\begin{figure}[h]
    \label{fig:data}
    \centering
    \includegraphics[width=0.45\textwidth]{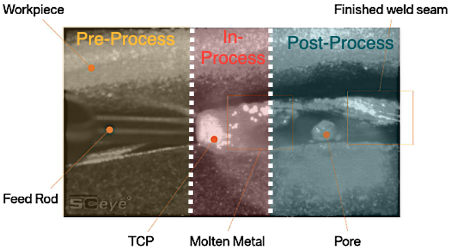}
    \caption{A Sample image data from the camera module at the LBW station. The components are marked.}
    \label{fig:data_sample}
\end{figure}

The image data for the various working scenarios are collected as the initial workflow. The Figure \ref{fig:data_sample} shows a sample data from the LBW unit having the failure class "Pore". A variance of data has been collected considering a combination of working conditions (angles, welding specifications, speed), materials etc. and a the initial dataset has been generated. 
Process external factors such as pre-existing surface impurities on the welding surface, cleanliness of protective glass on the camera module, exposure time of the camera module, digital filters used in the camera software, noise originated from the TCP of welding etc. also found to influence the quality of the welding as well as the quality of the data collected during the process. Due to the industrial nature of the use case and the uncontrollable involvement of the external factors, a diverse dataset consisting of the target object class "pore" has been collected from the welding stations for the model training. The collected initial dataset was consisting nearly 700 gray scale images with a frame size of $640\times320$ pixels. As a reasonable classification accuracy can be achieved even with a smaller sample sizes \cite{figueroa2012predicting}, the initial model developments have been performed using this original dataset. 

\section{Data Preparation}
\label{sec:data-preparation}
As the post-process of the welding gives more information on the prediction of the failures on the finished surface, the original image data has been reshaped to carry more information of the post-process, to enable the DL algorithms to learn data features efficiently. Hence the original image data has been cropped to a lower size of $300\times300$ pixels, matching the input size of the SOTA networks. Since the SOTA networks for transfer learning mostly accept three channel image as input, RGB images are generated from the original gray scale image through channel duplication. The images then further undergone various image enhancement and normalization methods as pre-processing. Later the pre-processed images are labelled with the target object classes using the open source labelImg \cite{tzutalin2015labelimg} tool in PASCAL VOC format for the TensorFlow implementation. 

\section{Modelling \& Fine-Tuning}
\label{sec:modelling}
As explained earlier in the methodology, the baseline models are developed as the initial step. This approach has been aimed at understanding the transfer learning performance of models on the original dataset. Taking insights from the background study, two SOTA object detection architectures, SSD Mobilenet and Faster-RCNN are chosen as baseline models. Faster-RCNN network performed with higher mAP values on COCO dataset compared to the SSD Mobilenet. After evaluating the performance of the baseline models in inference time, precision and accuracy based on COCO matrix, the fine tuning criteria are defined.      
\subsection{Baseline Modeling}
In a baseline model development, both SSD Mobilenet and Faster-RCNN are trained on the labelled original dataset. 
The model training are performed till the total loss values are converged to a minimum loss value, at the same time delivers reasonable results on validation data. The baseline model SSD Mobilenet V2 was supported with a SSD300 structure and the Faster-RCNN network used Inception V2 as backbone. Both the models are trained with a batch size of 32, till a reasonable lower loss value on training data and a higher mAP and mAR values on validation data, when evaluated with COCO matrix are obtained. After completing the training, the models are exported and evaluated against a common test dataset. The inference time of the models are estimated on a standard laptop (equipped with Intel Core i5-7300U CPU with 4 cores of 2.60Hz, 2.71 Hz and 8GB RAM) and the Faster-RCNN found to take $20\times$ more time compared to the SSD Mobilenet model for the inference on a single image input. Since both the models gave similar mAP and mAR values, due to the time constraints in the industrial application of the use case, the SSD Mobilenet is selected for the further fine tuning.  

\subsection{Model Fine-Tuning}
\label{sec:modelling_fine}
As the SSD Mobilenet network made faster inferences with similar performance scalars compared to the Faster-RCNN upon baseline modelling, considering the industrial application of the use case, the SSD Mobilenet has been used for the further research and fine-tuning. As the original dataset was quite small for generalization of the data through higher batch size training, dataset scaling through image augmentations found to be a feasible fine-tune approach. Hence various augmentation methods are applied on the original dataset using the \textit{imgaug} python package \cite{imgaug} and scaled datasets of $2\times-$, $4\times-$, $6\times-$ and $8\times-$ the original size are generated. Various augmentation methods such as: affine transformation (scaling \& translation), perspective transformations (horizontal \& vertical flips), image blurring (Gaussian blur), contrast changes, Gaussian Noise addition are applied on the original dataset. The rotation augmentation method is not applied on the data, as the bounding boxes on rotation augmentation undergoes shape transformations and gave incorrect labelled regions. A random combination of two of the augmentation methods are applied on each image and scaled datasets are generated. The Table \ref{tab:data_augmentation_compare} compares the object sizes (small, medium \& large) of the generated datasets against the original dataset. The majority of the object class in the training set was having medium size. 
\begin{table}[h]
\centering
\begin{tabular}{|l|c|c|c|}
\hline
\multicolumn{1}{|c|}{\textbf{\begin{tabular}[c]{@{}c@{}}Dataset\end{tabular}}} & \textbf{\begin{tabular}[c]{@{}c@{}}Small Pores\\ (pore\textless$32^2$\\px)\end{tabular}} & \textbf{\begin{tabular}[c]{@{}c@{}}Medium Pores\\ ($32^2$px\textless pore \textless \\$64^2$px)\end{tabular}} & \textbf{\begin{tabular}[c]{@{}c@{}}Large \\Pores\\ (pore\textgreater\\ $64^2$px)\end{tabular}} \\ \hline
\begin{tabular}[c]{@{}l@{}}Original\end{tabular}                             & 36                                                                                    & 401                                                                                                    & 131                                                                                      \\ \hline
\begin{tabular}[c]{@{}l@{}}$2\times\ $Scaled\end{tabular}               & 82                                                                                    & 776                                                                                                    & 267                                                                                      \\ \hline
\begin{tabular}[c]{@{}l@{}}$4\times\ $Scaled\end{tabular}                & 176                                                                                   & 1546                                                                                                   & 536                                                                                      \\ \hline
\begin{tabular}[c]{@{}l@{}}$6\times\ $Scaled\end{tabular}                & 389                                                                                   & 2274                                                                                                   & 691                                                                                      \\ \hline
\begin{tabular}[c]{@{}l@{}}$8\times\ $Scaled\end{tabular}                & 358                                                                                   & 3053                                                                                                   & 1061                                                                                     \\ \hline
\end{tabular}
\caption{The division of the object sizes in the generated training datasets.}
\label{tab:data_augmentation_compare}
\end{table} 

The SSD Mobilenet has later trained on the scaled datasets with higher batch sizes and the performance of the model in detecting the object classes are evaluated using the COCO matrix (against various IOUs and scale factors).
\section{Results}
\label{sec:results}

\begin{figure}[h]
    \centering
    \includegraphics[scale=0.6]{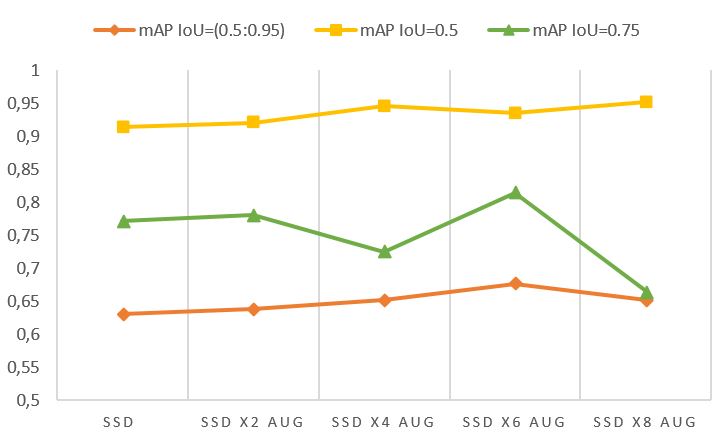}
    \caption{The mAP values of the models displayed in a single plot. The SSD Mobilenet model trained on the multiple scales of the original dataset have been compared against the mAP values with various IOUs. The model performances tend to increase with the data augmentation till a certain level and then reduces. The Y-axis shows the mAP.}
    \label{fig:map_summary}
\end{figure}

Keeping the precision (mAP) and recall (mAR) as the primary scalars of the measure of the object detection model, the performance of the exported models are evaluated on a common test dataset. The figure \ref{fig:map_summary} shows the precision (mAP) of the fine tuned models trained on the augmented datasets (four models denoted as : SSD $\times2$ Aug, SSD $\times4$ Aug, SSD $\times6$ Aug and SSD $\times8$ Aug) in comparison with the initial baseline model SSD Mobilenet trained on the original dataset (denoted as $SSD$). The baseline SSD Mobilenet model had a precision value of $0.630$, averaged over the 10 IoU values $\{0.5:0.95\}$. Upon data augmentation the average precision, calculated over the 10 IoU values, has been increased to $0.676$ at the $\times 6$ fold of augmentation. Through dataset scaling the overall precision has been increased by almost $4 \%$. But at higher level of augmentation, at $\times 8$ fold, the mAP reduced to a value of $0.652$.

\begin{figure}[h]
    \centering
    \includegraphics[scale=0.6]{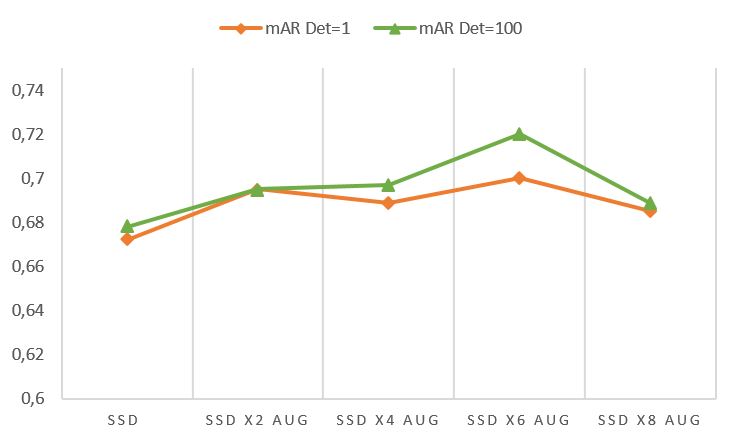}
    \caption{The mAR values of the models displayed in a single plot, comparing the recall values of the fine-tuned models with the baseline model. The models tend to improve their performances till a level of augmentation and then deteriorate. The Y-axis shows mAR value.}
    \label{fig:mar_summary}
\end{figure}

The figure \ref{fig:mar_summary} compares the recall (mAR) of the fine-tuned models to that of the initial baseline model. Upon synthetic data generation, at augmentation level of $\times 6$, the recall value has been increased from $0.678$ (baseline) to $0.720$ for the detection rate of 100. But at $\times8$ augmentation, the mAR value has been reduced to $0.689$.

\begin{figure}[h]
    \centering
    \includegraphics[scale=0.55]{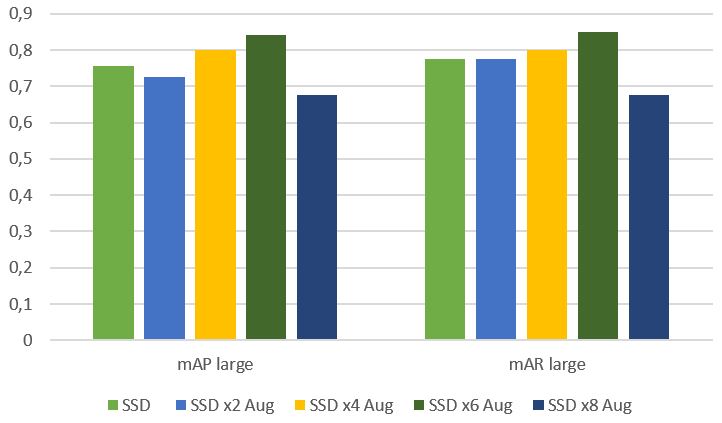}
    \caption{The mAP and mAR values of the developed models in detecting the larger pores. The precision and recall values of the models observed to increase along with the level of augmentation till a factor of $\times 6$. The Y-axis shows the values of mAP and mAR.}
    \label{fig:large_summary}
\end{figure}

As the research also focuses on the object sizes for the comparison, the object scaling matrices in the COCO are also considered for the performance analysis. On the image data, 40 px are equal to one millimeter on surface. The figure \ref{fig:large_summary} compares the precision and recall values of the developed models in detecting the large sized (pore\textgreater $64^2$px) object classes in the dataset. The baseline model had a mAP value of $0.756$ and a mAR value of $0.775$ in detecting the larger pores. Through augmentation the values of mAP has increased to $0.842$ and the mAR to $0.850$ (on $\times6$ augmentation). Upon further augmentation, at $\times8$ fold, the mAP and mAR values have observed to drop to $0.675$. The augmentation has improved the model performance in precision by $8.5\%$ and recall by $7.5\%$ in detecting larger pores.

\begin{figure}[h]
    \centering
    \includegraphics[scale=0.55]{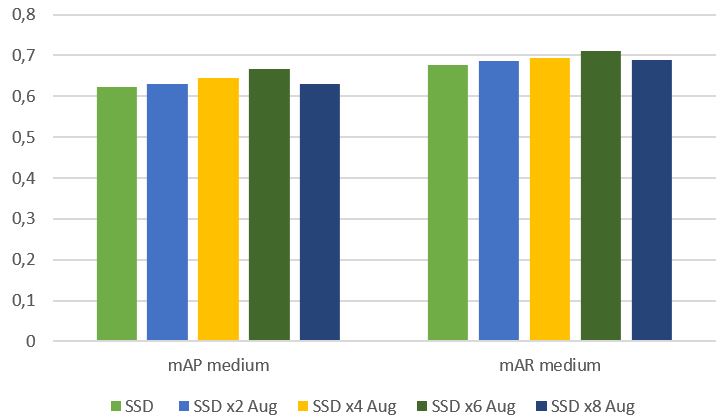}
    \caption{The mAP and mAR values of the developed models in detecting the medium sized pores. The best performing model has been identified at a scaling factor of $\times6$. The y-axis shows the values of mAP and mAR.}
    \label{fig:medium_summary}
\end{figure}

The figure \ref{fig:medium_summary} shows the mAP and mAR values of the models in detecting the medium sized pores ($32^2$px\textless pore \textless $64^2$px). As the datasets were having good share of medium sized object classes, the model behaviour can be compared to the effect of the dataset scaling in model performance. The baseline model after training on the original dataset delivered a precision of $0.624$ and a recall of $0.676$ for medium sized object classes on test data. Upon dataset scaling, the precision increased and reached a value of $0.666$ and the recall increased to a value of $0.712$ at $\times6$ fold. Upon further augmentation, at $\times 8$ fold, the precision as well the recall dropped to $0.630$ and $0.690$ respectively. The process of dataset scaling through augmentation increased the mAP in detecting medium sized objects by $4.2\%$ and mAR by $3.6\%$.

\begin{figure}[h]
    \centering
    \includegraphics[scale=0.550]{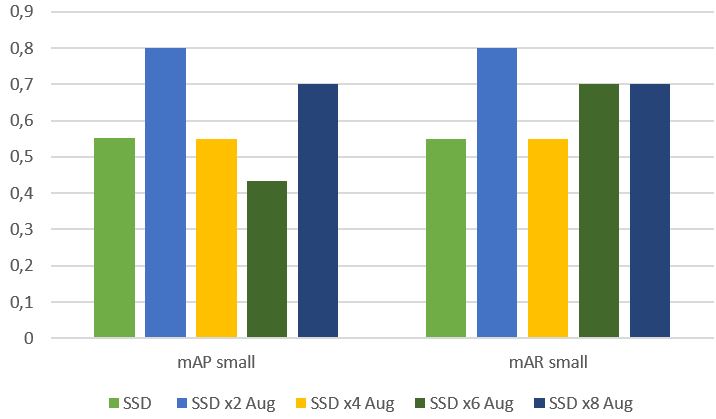}
    \caption{The mAP and mAR of the models in detecting the smaller sized pores. The best performing model has been identified at a scaling factor of $\times 2$. The Y-axis shows the percentage values of precision and recall. }
    \label{fig:small_summary}
\end{figure}

The figure \ref{fig:small_summary} shows the mAP and mAR values of the models in identifying smaller pores (pore\textless$32^2$px). As given in the table \ref{tab:data_augmentation_compare}, the initial dataset itself was having only a small share of pores of smaller size. The baseline SSD Mobilenet delivered a precision of $0.551$ and a recall of $0.550$ for smaller pore detection. On initial synthetic data generation, at $\times 2$ scaling, the precision and recall values showed a large improvement to a value of $0.800$.  But upon successive augmentation levels, the model tend to decrease the performance in both precision and recall values. As the possibilities of bias in the dataset, SSD Mobilenet network on the lightly augmented dataset ($\times 2$ fold), found to have superior performance compared to the other developed models.

\section{Conclusion}
\label{sec:conclusion}
This study gave more insights to the application of AI systems in manufacturing applications, proving its potential in quality control estimations. Through proper data understanding and data cleaning methods, it is feasible to develop, intelligent, data-driven DL models, conveniently with transfer learning. Inclusion of bias in the data, particularly through the data labelling, plays a vital role in a supervised data-driven model, badly influencing the model behaviours. Errors generated in the dataset through human mistakes also found drastically affecting the performance of algorithms. The dataset scaling through augmentation methods found to be an effective approach in performance improvement of the DL models, particularly in industrial applications where large datasets are difficult to obtain. 

Dataset scaling through augmentation is found to be an efficient method for the fine tuning of the models. The overall precision of the model has increased by $4.6\%$ and improved the precision in the PASCAL-VOC scalar (At IoU= $0.50$) of the baseline model from mAP of $0.914$ to a mAP of $0.951$ at $\times6$ fold.
The recall values found to increase gradually as the level of augmentation increases, till a certain fold of augmentation. At high augmentation levels the recall found to deteriorate. 
The performance of the models in detecting the large sized objects improved with the augmentation and reached a precision of $0.842$ and recall of $0.850$ at the $\times6$ fold of augmentation. Both precision and recall has been improved by $8\%$ through dataset scaling.
The precision and recall of the model in detecting the medium sized objects have been improved from marginally to a value of mAP of $0.666$ and mAR of $0.712$ at $\times6$ fold of augmentation. For both large and medium sized object both precision and recall values observed to drop with a high level of augmentation. 
The original dataset is observed to have only a marginal amount of data with smaller pores. An augmentation above $\times2$ fold was found to reduce model performance in the detection of smaller pores. This could be due to the introduction of bias in augmented dataset. The best precision and recall values are observed at light augmentation levels, with mAP and mAR values of $0.800$.

The models used in this research are implemented in the TensorFlow framework because of the popularity and larger open-source community support. But the latest SOTA systems such as Detectron 2 (from FAIR), YOLO V5 etc. offering higher performance in inference time, compared to these models. The high frequency demand of this use case offers higher potential for these models in the future researches. As these models are implemented in other frameworks having completely different use cases and objectives, and are still in the phase of active research, the performance of these models on industrial applications need to be verified. The developments in the high performance computer systems and numerous researches in the field of AI, particularly in DL, would definitely contribute more to efficient and advanced systems in the manufacturing industry in the near future.

\bibliography{IEEEabrv,main}

\end{document}

%% file: template/preamble.tex
\usepackage[a4paper,textwidth=18cm,textheight=24cm,top=2.85cm, bottom=2.85cm, left=1.5cm, right=1.5cm]{geometry}

\usepackage{icdp2009}

\makeatletter
\long\def\@makecaption#1#2{%
\vskip\abovecaptionskip
\sbox\@tempboxa{#1. #2}%
\ifdim \wd\@tempboxa >\hsize
#1. #2\par
\else
\global \@minipagefalse
\hb@xt@\hsize{\box\@tempboxa\hfil}%
\fi
\vskip\belowcaptionskip}
\makeatother